\documentclass{article}
\usepackage{ijcai24}

\usepackage{multirow}
\usepackage{pifont}
\usepackage{amssymb}
\usepackage{marvosym}

\usepackage{times}
\usepackage{soul}
\usepackage{url}
\usepackage[hidelinks]{hyperref}
\usepackage[utf8]{inputenc}
\usepackage[small]{caption}
\usepackage{graphicx}
\usepackage{amsmath}
\usepackage{amsthm}
\usepackage{booktabs}
\usepackage{algorithm}
\usepackage{algorithmic}
\usepackage[switch]{lineno}

\title{Detecting AI-Generated Sentences in Human-AI Collaborative Hybrid Texts: Challenges, Strategies, and Insights}

\author{
    Zijie Zeng$^{1}$\and
    Shiqi Liu$^{1,2}$\and
    Lele Sha$^{1}$\and
    Zhuang Li$^1$\and
    Kaixun Yang$^{1}$\\
    Sannyuya Liu$^{2}$\and
    Dragan Ga\v{s}evi\'{c}$^{1}$\And
    Guanliang Chen$^{1,}$\thanks{Corresponding author}
    \affiliations
    $^1$Centre for Learning Analytics, Monash University, Australia\\
    $^2$National Engineering Research Center of Educational Big Data, Central China Normal University, China\\
    \emails
    \{zijie.zeng, lele.sha, zhuang.li, kaixun.yang1, dragan.gasevic, guangliang.chen\}@monash.edu\\liushiqi@mails.ccnu.edu.cn, liusy027@ccnu.edu.cn
}

\begin{document}

\maketitle

\begin{abstract}

This study explores the challenge of sentence-level AI-generated text detection within human-AI collaborative hybrid texts (abbreviated as \textit{hybrid texts}). Existing studies of AI-generated text detection for hybrid texts often rely on synthetic datasets. These typically involve hybrid texts with a limited number of boundaries, e.g., single-boundary hybrid texts that begin with human-written content and end with machine-generated continuations. We contend that studies of detecting AI-generated content within hybrid texts should cover different types of hybrid texts generated in realistic settings to better inform real-world applications. Therefore, our study utilizes the CoAuthor dataset, which includes diverse, realistic hybrid texts generated through the collaboration between human writers and an intelligent writing system in multi-turn interactions. We adopt a two-step, segmentation-based pipeline: (\romannumeral1) detect segments within a given hybrid text where each segment contains sentences of consistent authorship, and (\romannumeral2) classify the authorship of each identified segment. Our empirical findings highlight (1) detecting AI-generated sentences in hybrid texts is overall a challenging task because (1.1) human writers' selecting and even editing AI-generated sentences based on personal preferences adds difficulty in identifying the authorship of segments; (1.2) the frequent change of authorship between neighboring sentences within the hybrid text creates difficulties for segment detectors in identifying authorship-consistent segments; (1.3) the short length of text segments within hybrid texts provides limited stylistic cues for reliable authorship determination; (2) before embarking on the detection process, it is beneficial to assess the average length of segments within the hybrid text. This assessment aids in deciding whether (2.1) to employ a text segmentation-based strategy for hybrid texts with longer segments, or (2.2) to adopt a direct sentence-by-sentence classification strategy for those with shorter segments.

\end{abstract}

\section{Introduction}\label{sec:intro}

AI-generated text detection refers to the process of identifying the text generated by AI systems. The significance of this task lies in its dual role as both a deterrent and a defense against the potential misuse of modern, powerful generative AI technologies, particularly in educational contexts. According to \cite{ma2023abstract,zellers2019defending}, contemporary advanced language models are capable of producing deceptive content designed to manipulate public perception. This includes the creation of fictitious news \cite{zellers2019defending}, counterfeit application reviews \cite{martens2019towards}, and fabricated social media posts \cite{fagni2021tweepfake}. Additionally, education practitioners have expressed concerns about the growing popularity of the use of ChatGPT (a generative AI system developed by OpenAI) among students. They emphasize that over-reliance on ChatGPT could hinder the development of students' writing and critical thinking skills \cite{ma2023abstract,mitchell2023detectgpt} and, in the worst cases, lead to breaches of academic integrity \cite{currie2023academic,sun2023chatgpt}.

Existing studies \cite{Koike-OUTFOX-2024,hu2024bad,ma2023abstract,mitchell2023detectgpt,clark2021all,jawahar2020automatic,martens2019towards} have primarily focused on detecting AI-generated text at the granularity of paragraphs or documents, operating under the assumption that the text under scrutiny is either entirely AI-generated or not at all. However, the framing of the detection task as a binary classification problem at the document level has recently been challenged by \cite{dugan2023real}, who have observed the trends in human-AI collaborative writing \cite{buschek2021impact,lee2022coauthor}. They noted that a piece of text (e.g., a document) could contain both human-written and AI-generated content. This observation underscores the need for detecting AI-generated text at a finer granularity level (e.g., at the sentence level instead of the document level). Although pilot studies \cite{seqxgpt_2023,zeng2023towards} have been conducted to detect AI-generated content within hybrid texts, we are \textbf{concerned} that the strategies these studies used to synthesize hybrid texts are \textbf{overly simplistic and lack variation}. For example, a simple practice of synthesizing a hybrid text (as adopted by \cite{seqxgpt_2023,zeng2023towards}) involves using a piece of human-written text as a starting point, and then instructing a generative AI tool (e.g., ChatGPT) to write a continuation based on the given human-written beginning text, resulting in a synthesized hybrid text with \textbf{only one boundary}. Behind this concern lies our doubt regarding \textit{whether a detection strategy or approach that works well on relatively simple hybrid texts (e.g., those with fewer boundaries such as the single-boundary ones) can be equally effective on complex hybrid texts, e.g., hybrid texts with \textbf{up to dozens of boundaries} that are generated through multi-turn interactions between human writers and intellectual writing assistants}. To investigate this, we posit that a comprehensive study of AI-generated text detection from hybrid texts should be grounded on a dataset that encompasses hybrid texts generated with varying human-AI interaction styles and strategies. Therefore, in this study, we conducted further investigation to detect AI-generated text within hybrid texts. It is noteworthy that these hybrid texts originate from the CoAuthor dataset \cite{lee2022coauthor} and are collaboratively created by human writers and intelligent writing assistants through a series of interactions in real-world environments. We formally define our \textbf{R}esearch \textbf{Q}uestion as:

\begin{itemize}
	\item To what extent can AI-generated sentences be automatically detected in hybrid texts co-created by human writers and intelligent writing assistants?
\end{itemize}
Recognizing the fact that a hybrid text can be divided into segments, with each segment consisting of several sentences that share the same authorship, we use a two-step segmentation-based pipeline \cite{tepper2012statistical,xia2023sequence} (i.e., \textbf{segment detection} followed by \textbf{segment classification}) for sentence-level AI-generated text detection from hybrid texts. However, as noted in \cite{zeng2023towards}, as the number of boundaries increases, segment detection approaches tend to become less accurate in identifying boundaries between human-written and machine-generated segments. Thus, we also introduce a naive segmentation strategy, treating each sentence as an individual segment, to facilitate comparisons between text segmentation-based strategies and the straightforward sentence-by-sentence classification strategy. Through extensive empirical experiments, we have summarized the following main findings: (1) detecting AI-generated sentences within hybrid texts is overall a challenging task because (1.1) the complex interplay between human and AI writing systems, where humans could select and even edit AI-generated sentences based on personal preferences, adds difficulty in identifying the authorship of segments; (1.2) the frequent change of authorship between neighboring sentences within the hybrid text creates difficulties for segment detectors in accurately identifying authorship-consistent segments, thereby hampering the downstream authorship identification task; (1.3) the short length of text segments offering limited stylistic information for accurate authorship identification; (2) with a sufficiently high-performing segment detection model and segment classification model, the two-step segmentation-based pipeline solution may perform better than approaches that jointly learn these two tasks; (3) Before starting the detection task, assessing (through prior knowledge or other machine learning techniques) the \textbf{average segment length} of hybrid texts is useful, as it can influence the strategy of either: (3.1) using the text segmentation-based strategy for hybrid text with longer segments; or (3.2) opting for the direct sentence-by-sentence classification strategy for hybrid text with shorter segments\footnote{The concept of \textbf{average segment length} can be illustrated with the following two hybrid texts $T = <h_{1}, h_{2}, h_{3}, m_{4}>$ and $R = <h_{5}, m_{6}, h_{7}, m_{8}>$, where $h_i$ and $m_i$ are \textbf{h}uman-written sentence and \textbf{m}achine-generated sentence, respectively. The average segment length $l_T$ for $T$ can be calculated in a weighted manner as $l_T = (3/4 \times 3) + (1/4 \times 1) = 2.5$, while $l_R=(1/4+1/4+1/4+1/4)\times 1=1<2.5$, i.e., \textbf{a shorter average length reflects the more frequent change of authorship between neighboring sentences}. Note that we assume each sentence has a length of $1$.}.

\section{Related Work}
\smallskip
\noindent\textbf{Al-Generated Text Detection.}
As modern LLMs' generative capabilities strengthen, distinguishing between AI-generated texts and human-written texts becomes increasingly challenging \cite{ma2023abstract}. This may lead to the misuse of generative AI in scenarios where it should not be used, such as students employing it for writing assignments, which can result in academic misconduct \cite{mitchell2023detectgpt}. These potential risks emphasize the necessity for effective detection of AI-generated text. Existing studies \cite{Koike-OUTFOX-2024,hu2024bad,he2023DeBERTav,mitchell2023detectgpt,pagnoni2022,synscipass} have been criticized for their narrow focus on document-level detection and the assumption that a document is either entirely AI-generated or not. This point is noted in \cite{dugan2023real}, which suggests that AI-generated text detection should be conducted with finer granularity to adapt to the increasingly popular trend of human-AI collaborative writing. Specifically, \cite{zeng2023towards} introduced a boundary detector for detecting boundaries between human-written sentences and AI-generated sentences. Meanwhile, \cite{seqxgpt_2023} adopted a more straightforward approach to detection, namely, identifying the exact authorship of each sentence within the hybrid text. They proposed a method called SeqXGPT to extract word-wise log probability lists and use a combination of CNN and Transformer for sentence-level authorship detection within hybrid texts.

\smallskip
\noindent\textbf{Text Segmentation.}
Text segmentation aims to discover the latent structure of a document by dividing the text into segments with different labels \cite{text-seg-2023-lessons,bai2023SegFormer,lo2021transformer,xia2023sequence} (e.g., segmentation based on different topics). The broad spectrum of text segmentation studies can be further divided into two categories: (1) those focusing solely on boundary detection for segmentation \cite{zeng2023towards,lukasik2020text,text-seg-2023-lessons,yu2023improving,xing2020improving,li2022human,somasundaran2020two,koshorek2018text}; and (2) those aiming at both segment detection and subsequent segment classification \cite{bai2023SegFormer,lo2021transformer,xia2023sequence,gong2022tipster,tepper2012statistical}. Among the many possible approaches for segmentation, we have selected three approaches with open-source code for our segment detection module: (\romannumeral1) \texttt{TriBERT} \cite{zeng2023towards}, because it was specifically developed for distinguishing AI-generated segments and human-written segments within hybrid texts; (\romannumeral2) \texttt{Transformer$^2$} \cite{lo2021transformer}, due to its widespread adoption as a baseline in existing text segmentation studies \cite{xia2022dialogue,xia2023sequence,bai2023SegFormer,yu2023improving}; and (\romannumeral3) \texttt{SegFormer} \cite{bai2023SegFormer}, chosen as a representative of the state-of-the-art (also one of the latest) text segmentation approaches.

\smallskip
\noindent\textbf{Human-AI Collaborative Writing and Datasets.}
Modern large language models have made human-AI collaborative writing increasingly convenient. Existing studies show that the suggestions provided by intelligent writing assistants have evolved from phrase-level \cite{buschek2021impact} to sentence-level \cite{lee2022coauthor}. Furthermore, these assistants seem poised for further improvement, especially with the release of GPT-4. Pilot studies \cite{dugan2023real,zeng2023towards,seqxgpt_2023} have been conducted to address AI-generated text detection within hybrid text data at a granularity finer than the document level. However, these studies were mainly conducted on synthetic and simplistic datasets due to their low cost and ease of acquisition. For instance, \cite{seqxgpt_2023} instructed the GPT-3.5-turbo API to generate continuations on given human-written texts. \cite{zeng2023towards} goes further by designing a set of six prompting templates that enable ChatGPT to generate multi-boundary hybrid texts (no more than three boundaries) from incomplete student-written essays. For each of these original essays, specific segments were removed and substituted with new sentences generated by ChatGPT. However, these synthetic texts are still overly simplistic compared to the hybrid texts generated through multi-turn human-AI interactions in realistic scenarios, which could involve up to dozens of boundaries. We contend that to guide and enlighten real-world practices, AI-generated text detection studies should not rely solely on synthetic datasets. Therefore, our study aims at the CoAuthor dataset \cite{lee2022coauthor}, which includes diverse, realistic hybrid texts generated through human-AI interactions and can serve as an ideal benchmark for exploring AI-generated text detection in hybrid texts.

\section{Methods}
\subsection{Dataset and Task}
\smallskip
\noindent\textbf{Dataset.}
We selected the CoAuthor Dataset \cite{lee2022coauthor}, which consists of texts generated through multi-turn interactions between human writers and a GPT-3-based intelligent writing assistant. The dataset includes $1,445$ essays from $63$ writers who interacted with the writing assistant across multiple writing sessions. Writers were given a prompt text as a starter and could either write independently or request sentence suggestions from the assistant. These suggestions could be accepted (with or without edits) or rejected, and this process continued until an essay was completed\footnote{See https://coauthor.stanford.edu/ for more about CoAuthor.}.

\smallskip
\noindent\textbf{Task.}
For a hybrid text $H$ comprising $n$ sentences, represented as a sequence of sentences $\{s_1,s_2,...,s_n\}$, an algorithm is required to predict a list of labels $\{l_1,l_2,...,l_n\}$ for $H$. Each label $l_i$ corresponds to the predicted authorship for sentence $s_i$, e.g., in CoAuthor dataset, $l_i$ could be \texttt{AI-generated}, \texttt{Human-written}, or \texttt{Human-AI Collaborative}.

\subsection{Approaches}\label{sec:models}
We adopt a two-step segmentation-based pipeline to address the sentence-level AI-generated text detection from hybrid texts, involving segment detection followed by segment classification \cite{tepper2012statistical,xia2023sequence}. This pipeline approach is based on our observation that a hybrid text can be divided into segments, with each segment consisting of several sentences that share the same authorship. We begin by introducing the \textbf{segment detection module}:

\begin{itemize} 
    \item[$\bullet$] \textbf{Perfect Segment Detector}: This approach serves as an ideal segment detector, predicting all boundaries with $100\%$ accuracy.
    \item[$\bullet$] \textbf{Naive Segment Detector}: This approach treats each sentence in the hybrid text as a separate segment, which is essentially equivalent to performing \textbf{direct sentence-by-sentence classification} over the hybrid text.
    \item[$\bullet$] \textbf{TriBERT}: \texttt{TriBERT} \cite{zeng2023towards} is designed to identify transitions between segments of differing authorship (e.g., human vs. AI-generated text). It measures distances between adjacent segments, pinpointing transitions where this distance is greatest. However, it only ranks boundary candidates by their probabilities of being true boundaries. Therefore, we manually selected the top $p$ boundaries from the sorted candidate list with $p\in \{20\%,40\%,60\%,80\%\}$\footnote{\texttt{TriBERT} with $p=100\%$ is \textbf{equivalent to} the \texttt{Naive} segment detector, i.e., the \textbf{sentence-by-sentence classification} strategy.}.
    \item[$\bullet$] \textbf{SegFormer}: \texttt{SegFormer} \cite{bai2023SegFormer} is a topic segmentation approach that utilizes unidirectional attention blocks for enhanced sentence modeling. It incorporates a specialized context aggregator and leverages topic classification loss to reduce noise and enhance information processing. Note that SegFormer is capable of not only identifying boundaries between segments but also providing a predicted label for each segment.
    \item[$\bullet$] \textbf{Transformer$^2$} \cite{lo2021transformer}: This approach merges bottom-level pre-trained transformers for sentence embedding with an upper-level transformer to identify segments and predict their labels (i.e., authorship).
	
\end{itemize}
Then we present the approaches for the \textbf{segment classification module} as follows:
\begin{itemize}
	\item[$\bullet$] \textbf{BERT}:  This method utilizes the BERT-base \cite{kenton2019BERT} model to encode the textual input into an embedding. This embedding is then fed into a dense classification layer for label prediction.
	\item[$\bullet$] \textbf{SeqXGPT}: \texttt{SeqXGPT} \cite{seqxgpt_2023} utilizes the log probability lists generated by transparent LLMs as its primary features. Considering that these features mirror the wave patterns observed in speech processing and surpass the analytical capabilities of LLMs, it further integrates convolutional and self-attention networks into its core architecture to process these features.
	\item[$\bullet$] \textbf{DeBERTa-v3}:  \texttt{DeBERTa-v3} \cite{he2023DeBERTav} is a recently developed language model that has set new benchmarks in several natural language processing (NLP) tasks. This advanced version builds on its predecessor, \texttt{DeBERTa} \cite{he2020DeBERTa}, by implementing an enhanced pre-training task with improved sample efficiency called replaced token detection (RTD), instead of the traditional masked language modeling task.
	\item[$\bullet$] \textbf{DistilBERT}:  This method resembles the above method, \texttt{BERT}, but it adopts an alternative encoder, namely DistilBERT \cite{sanh2019distilBERT}, which is a distilled, efficient version of BERT.
	
	\item[$\bullet$] \textbf{RoBERTa}:  This method resembles the above method, \texttt{BERT}, but it adopts RoBERTa \cite{liu2019roBERTa} as the encoder, which can be viewed as an optimized version of BERT, with changes primarily in the training regime and data size.
	
	\item[$\bullet$] \textbf{GPT-3.5 (Fine-tuned)}:  We first format our training data as follows: Input: $<$TEXT SEGMENT$>$, Output: $<$LABEL$>$. Then, we fine-tune the OpenAI GPT-3.5-Turbo-1106 API using the above-formatted training data and obtain our AI-generated text classifier.
    
	\item[$\bullet$] \textbf{GPT-2}:  This method adopts the GPT-2 \cite{radford2019language} model with a classification head on top of the hidden states of the last token for classification.
	
	\item[$\bullet$]\textbf{BERT (Token)}, \textbf{DistilBERT (Token)}, and \textbf{RoBERTa (Token)}: 
	This series of methods classifies each token in a text segment, then determines the segment's label based on token-level predictions. If all tokens are predicted as \texttt{AI-generated} or \texttt{human-written}, the segment is labeled accordingly. If both types are present, the segment is considered as \texttt{human-AI collaborative}. We use BERT, DistilBERT, and RoBERTa as encoders for token embeddings, followed by a classification layer for token-level authorship prediction. These methods are termed \texttt{BERT (Token)}, \texttt{DistilBERT (Token)}, and \texttt{RoBERTa (Token)}.   
	\item[$\bullet$] \textbf{Transformer$^2$} and \textbf{SegFormer}: As mentioned above, these two methods can identify boundaries for text segmentation as well as predict labels for the identified segments. Therefore, we also consider \texttt{Transformer$^2$} and \texttt{SegFormer} as segment classification models.
\end{itemize}

\subsection{Experiment Setting}

\smallskip
\noindent\textbf{Data Splitting.}
We randomly divided the $1,445$ hybrid texts in the CoAuthor dataset into training, validation, and testing sets in a ratio of 70\%:15\%:15\%. Note that the above split datasets were used for the training, validation, and testing processes of all models, including segment detection models and segment classification models.

\smallskip
\noindent\textbf{Evaluation Metric.}
We observed that accurately detecting segments within a hybrid text does not ensure the precise identification of the authorship for these detected segments. For example, in a hybrid text comprising four sentences $<s_1, s_2, s_3, s_4>$, if a segment detector correctly identifies the boundary between sentences $s_2$ and $s_3$, it remains ambiguous whether the segment $<s_1, s_2>$ (or $<s_3, s_4>$) is AI-generated or human-written. \textbf{The segment detector merely indicates a transition in authorship}, without specifying which segment is AI-generated. Consequently, we opted for the \textbf{Kappa} score as our evaluation metric, which measures the inter-rater agreement between lists of predicted and actual authorship labels, rather than the $P_k$ metric, commonly adopted for evaluating text segmentation performance \cite{beeferman1999statistical,bai2023SegFormer,lo2021transformer}.

\smallskip
\noindent\textbf{Model Training.} As training details vary across different models, we itemize them as follows:
\begin{itemize}
	\item [$\bullet$] \textbf{TriBERT}: We adhere to its default setting \cite{zeng2023towards} to evaluate the model's performance on the validation set after each training epoch\footnote{For this approach, an epoch is defined as the completion of training on $n$ (we set $n=5000$) samples, where a sample is a triplet of sentences, including a pair of sentences consistent in authorship and a third sentence whose authorship differs from that of the pair.}. If the validation performance at epoch $t+1$ is not as good as at epoch $t$, we stop the training and use the best model (saved at epoch $t$) as the segment detector for the test stage.
	\item [$\bullet$] \textbf{Naive Segment Detector and Perfect Segment Detector}: These two models require no training process. The first model treats each sentence as a separate segment, while the second model directly splits the hybrid texts into segments using the ground-truth boundaries.
	
	\item [$\bullet$] \textbf{SegFormer and Transformer$^2$}: These two approaches consider the whole piece of hybrid text as the minimum input and define an epoch as a single iteration of training across all hybrid texts in the training set. 
	
	\item [$\bullet$] \textbf{Segment Classification Models}: For all segment classification models, a training sample refers to a single sentence. During the test phase, the authorship of a segment is determined based on the majority authorship of its sentences, as predicted from a list containing the authorship for each sentence within the segment.
	
\end{itemize}

\smallskip
\noindent\textbf{Important Parameters and Other Details.}
For all models, the learning rate was selected from $\{2e-5, 5e-5, 1e-4\}$. We used a small batch size of $4$ for \texttt{SegFormer} due to its higher memory requirements. For other approaches, we consistently adopted a batch size of $32$. All tests were conducted using an NVIDIA Tesla A40 GPU equipped with $48$ GB of RAM. For details of other model-specific parameters, please refer to our codes available on GitHub\footnote{https://github.com/douglashiwo/AISentenceDetection}.

\section{Results}\label{sec:results}
\begin{table*}[!htb]
	\begin{center}
		
		\resizebox{0.86\textwidth}{!}{
			
			\begin{tabular}{l|ccccc|c|c|c|c} 
				\toprule
				\multicolumn{10}{c}{\textbf{G0: The overall performance on test set}} \\ 
				\midrule
				\multirow{2}{*}{Method} & \multicolumn{5}{c|}{TriBERT} & \multirow{2}{*}{\textbf{C6}: SegFormer} & \multirow{2}{*}{\textbf{C7}: Transformer$^2$} & \multirow{2}{*}{\textbf{C8}: Perfect} & \multirow{2}{*}{\textbf{C9}: Mean} \\ 
				\cline{2-6}
				& \textbf{C1}: top20\% &\textbf{C2}: top40\% &\textbf{C3}: top60\% &\textbf{C4}: top80\% &\textbf{C5}: Naive & & & & \\ 
				\midrule
				\textbf{R1}: DeBERTa-v3 & 0.2246 & 0.2913 & 0.3391 & 0.3754 & \textbf{0.4002} & 0.3176 & 0.2931 & 0.5166 & 0.3447 \\
				\textbf{R2}: BERT & 0.2211 & 0.2748 & 0.3160 & 0.3528 & \textbf{0.3748} & 0.3015 & 0.2885 & 0.4960 & 0.3282 \\
				\textbf{R3}: SeqXGPT & 0.1686 & 0.2177 & 0.2627 & 0.3120 & \textbf{0.3522} & 0.2267 & 0.2116 & 0.3945 & 0.2683 \\
				\textbf{R4}: RoBERTa & 0.2262 & 0.2726 & 0.3099 & 0.3424 & \textbf{0.3607} & 0.3025 & 0.2656 & 0.4718 & 0.3190 \\
				\textbf{R5}: DistilBERT & 0.2157 & 0.2651 & 0.3036 & 0.3430 & \textbf{0.3612} & 0.2943 & 0.2694 & 0.4649 & 0.3147 \\
				\textbf{R6}: GPT-3.5 (Fine-tuned) & 0.2297 & 0.2316 & 0.2508 & 0.2650 & 0.2707 & \textbf{0.2747} & 0.2592 & 0.4492 & 0.2789 \\
				\textbf{R7}: BERT (Token) & 0.2053 & 0.2345 & 0.2529 & 0.2701 & 0.2651 & \textbf{0.2845} & 0.2307 & 0.3922 & 0.2669 \\
				\textbf{R8}: RoBERTa (Token) & 0.1665 & 0.1963 & 0.2152 & 0.2351 & 0.2285 & \textbf{0.2390} & 0.1899 & 0.3036 & 0.2218 \\
				\textbf{R9}: DistilBERT (Token) & 0.1924 & 0.2218 & 0.2432 & 0.2544 & 0.2491 & \textbf{0.2734} & 0.2185 & 0.3601 & 0.2516 \\
				\textbf{R10}: GPT2 & 0.1293 & 0.1665 & 0.1899 & \textbf{0.2090} & 0.2071 & 0.1905 & 0.1578 & 0.2473 & 0.1872 \\ 
				\midrule
				\textbf{R11}: Mean & 0.1979 & 0.2372 & 0.2683 & 0.2959 & \textbf{0.3070} & 0.2705 & 0.2384 & 0.4096 & --- \\
				\midrule
				\textbf{R12}: SegFormer & \multicolumn{9}{c}{\textbf{0.3180}} \\
				\textbf{R13}: Transformer$^2$ & \multicolumn{9}{c}{\textbf{0.2519}} \\ 
				\bottomrule
				\multicolumn{10}{c}{\textbf{G1: The performance on the group of test samples with average segment length $l<= 3$} } \\ 
				\midrule
				\multirow{2}{*}{Method} & \multicolumn{5}{c|}{TriBERT} & \multirow{2}{*}{\textbf{C6}: SegFormer} & \multirow{2}{*}{\textbf{C7}: Transformer$^2$} & \multirow{2}{*}{\textbf{C8}: Perfect} & \multirow{2}{*}{\textbf{C9}: Mean} \\ 
				\cline{2-6}
				&\textbf{C1}: top20\% &\textbf{C2}: top40\% &\textbf{C3}: top60\% &\textbf{C4}: top80\% &\textbf{C5}: Naive & & & & \\ 
				\midrule
				\textbf{R1}: DeBERTa-v3 & 0.1393 & 0.2198 & 0.2827 & 0.3431 & \textbf{0.3847} & 0.2531 & 0.2309 & 0.4513 & 0.2881 \\
				\textbf{R2}: BERT & 0.1409 & 0.2053 & 0.2644 & 0.3141 & \textbf{0.3524} & 0.2426 & 0.2197 & 0.4164 & 0.2695 \\
				\textbf{R3}: SeqXGPT & 0.0930 & 0.1464 & 0.2015 & 0.2666 & \textbf{0.3182} & 0.1795 & 0.1613 & 0.3461 & 0.2141 \\
				\textbf{R4}: RoBERTa & 0.1509 & 0.2179 & 0.2724 & 0.3209 & \textbf{0.3582} & 0.2564 & 0.2126 & 0.4141 & 0.2754 \\
				\textbf{R5}: DistilBERT & 0.1324 & 0.1960 & 0.2517 & 0.3016 & \textbf{0.3383} & 0.2332 & 0.1984 & 0.3871 & 0.2548 \\
				\textbf{R6}: GPT-3.5 (Fine-tuned) & 0.1471 & 0.1774 & 0.2072 & 0.2402 & \textbf{0.2612} & 0.2018 & 0.1820 & 0.3531 & 0.2213 \\
				\textbf{R7}: BERT (Token) & 0.1332 & 0.1883 & 0.2254 & 0.2620 & \textbf{0.2712} & 0.2316 & 0.1851 & 0.3498 & 0.2308 \\
				\textbf{R8}: RoBERTa (Token) & 0.1123 & 0.1582 & 0.1943 & 0.2337 & \textbf{0.2395} & 0.2050 & 0.1671 & 0.2958 & 0.2007 \\
				\textbf{R9}: DistilBERT (Token) & 0.1278 & 0.1783 & 0.2089 & 0.2404 & \textbf{0.2512} & 0.2118 & 0.1760 & 0.3193 & 0.2142 \\
				\textbf{R10}: GPT2 & 0.1067 & 0.1476 & 0.1813 & 0.2172 & \textbf{0.2256} & 0.1824 & 0.1557 & 0.2633 & 0.1850 \\ 
				\midrule
				\textbf{R11}: Mean & 0.1284 & 0.1835 & 0.2290 & 0.2740 & \textbf{0.3001} & 0.2197 & 0.1889 & 0.3596 & --- \\
				\midrule
				\textbf{R12}: SegFormer & \multicolumn{9}{c}{\textbf{0.2821}} \\
				\textbf{R13}: Transformer$^2$ & \multicolumn{9}{c}{\textbf{0.1939}} \\ 
				\bottomrule
				\multicolumn{10}{c}{\textbf{G2: The performance on the group of test samples with average segment length $l\in (3, 6]$}} \\ 
				\midrule
				\multirow{2}{*}{Method} & \multicolumn{5}{c|}{TriBERT} & \multirow{2}{*}{\textbf{C6}:SegFormer} & \multirow{2}{*}{\textbf{C7}: Transformer$^2$} & \multirow{2}{*}{\textbf{C8}: Perfect} & \multirow{2}{*}{\textbf{C9}: Mean} \\ 
				\cline{2-6}
				&\textbf{C1}: top20\% &\textbf{C2}: top40\% &\textbf{C3}: top60\% &\textbf{C4}: top80\% &\textbf{C5}: Naive & & & & \\ 
				\midrule
				\textbf{R1}: DeBERTa-v3 & 0.2447 & 0.3218 & 0.3674 & 0.3861 & \textbf{0.3989} & 0.3452 & 0.3016 & 0.5181 & 0.3605 \\
				\textbf{R2}: BERT & 0.2325 & 0.3036 & 0.3436 & 0.3793 & \textbf{0.3914} & 0.3206 & 0.3048 & 0.5225 & 0.3498 \\
				\textbf{R3}: SeqXGPT & 0.2030 & 0.2387 & 0.2834 & 0.3079 & \textbf{0.3291} & 0.2289 & 0.2113 & 0.4014 & 0.2755 \\
				\textbf{R4}: RoBERTa & 0.2465 & 0.3002 & 0.3364 & 0.3595 & \textbf{0.3634} & 0.3184 & 0.2631 & 0.4858 & 0.3342 \\
				\textbf{R5}: DistilBERT & 0.2371 & 0.3008 & 0.3333 & 0.3770 & \textbf{0.3831} & 0.3280 & 0.3003 & 0.4984 & 0.3448 \\
				\textbf{R6}: GPT-3.5 (Fine-tuned) & 0.2187 & 0.2086 & 0.2548 & 0.2759 & 0.2655 & \textbf{0.2785} & 0.2613 & 0.4400 & 0.2754 \\
				\textbf{R7}: BERT (Token) & 0.2304 & 0.2566 & 0.2815 & 0.2914 & 0.2845 & \textbf{0.3119} & 0.2466 & 0.4128 & 0.2895 \\
				\textbf{R8}: RoBERTa (Token) & 0.1989 & 0.2194 & 0.2389 & 0.2625 & 0.2531 & \textbf{0.2716} & 0.2208 & 0.3569 & 0.2528 \\
				\textbf{R9}: DistilBERT (Token) & 0.2246 & 0.2525 & 0.2822 & 0.2921 & 0.2828 & \textbf{0.3132} & 0.2564 & 0.4010 & 0.2881 \\
				\textbf{R10}: GPT2 & 0.1394 & 0.1768 & 0.2043 & \textbf{0.2184} & 0.2153 & 0.2098 & 0.1531 & 0.2649 & 0.1978 \\ 
				\midrule
				\textbf{R11}: Mean & 0.2176 & 0.2579 & 0.2926 & 0.3150 & \textbf{0.3167} & 0.2926 & 0.2519 & 0.4302 & ---\\
				\midrule
				\textbf{R12}: SegFormer & \multicolumn{9}{c}{\textbf{0.3235}} \\
				\textbf{R13}: Transformer$^2$ & \multicolumn{9}{c}{\textbf{0.2848}} \\ 
				\bottomrule
				\multicolumn{10}{c}{\textbf{G3: The performance on the group of test samples with average segment length $l>6$}} \\ 
				\midrule
				\multirow{2}{*}{Method} & \multicolumn{5}{c|}{TriBERT} & \multirow{2}{*}{\textbf{C6}: SegFormer} & \multirow{2}{*}{\textbf{C7}: Transformer$^2$} & \multirow{2}{*}{\textbf{C8}: Perfect} & \multirow{2}{*}{\textbf{C9}: Mean} \\ 
				\cline{2-6}
				&\textbf{C1}: top20\% &\textbf{C2}: top40\% &\textbf{C3}: top60\% &\textbf{C4}: top80\% &\textbf{C5}: Naive & & & & \\ 
				\midrule
				\textbf{R1}: DeBERTa-v3 & \textbf{0.4038} & 0.3965 & 0.3989 & 0.3863 & 0.3827 & 0.3938 & 0.3948 & 0.6486 & 0.4257 \\
				\textbf{R2}: BERT & \textbf{0.3834} & 0.3702 & 0.3620 & 0.3584 & 0.3469 & 0.3584 & 0.3811 & 0.6253 & 0.3982 \\
				\textbf{R3}: SeqXGPT & 0.1395 & 0.2140 & 0.2362 & 0.2719 & \textbf{0.3017} & 0.1451 & 0.1660 & 0.3479 & 0.2278 \\
				\textbf{R4}: RoBERTa & \textbf{0.3507} & 0.3277 & 0.3232 & 0.3197 & 0.3121 & 0.3374 & 0.3338 & 0.5471 & 0.3565 \\
				\textbf{R5}: DistilBERT & \textbf{0.3838} & 0.3561 & 0.3516 & 0.3458 & 0.3298 & 0.3416 & 0.3557 & 0.5742 & 0.3798 \\
				\textbf{R6}: GPT-3.5 (Fine-tuned) & \textbf{0.3974} & 0.3383 & 0.2870 & 0.2527 & 0.2428 & 0.3752 & 0.3703 & 0.6691 & 0.3666 \\
				\textbf{R7}: BERT (Token) & 0.2276 & 0.2249 & 0.2125 & 0.2086 & 0.1865 & \textbf{0.2666} & 0.2075 & 0.3777 & 0.2390 \\
				\textbf{R8}: RoBERTa (Token) & 0.1514 & 0.1795 & 0.1734 & 0.1620 & 0.1474 & \textbf{0.2040} & 0.1345 & 0.2128 & 0.1706 \\
				\textbf{R9}: DistilBERT (Token) & 0.2097 & 0.2118 & 0.2046 & 0.1867 & 0.1676 & \textbf{0.2632} & 0.1773 & 0.3248 & 0.2182 \\
				\textbf{R10}: GPT2 & 0.1054 & 0.1382 & 0.1465 & \textbf{0.1481} & 0.1368 & 0.1396 & 0.1204 & 0.1732 & 0.1385 \\ 
				\midrule
				\textbf{R11}: Mean & 0.2753 & 0.2757 & 0.2696 & 0.2640 & 0.2554 & \textbf{0.2825} & 0.2641 & 0.4501 & --- \\
				\midrule
				\textbf{R12}: SegFormer & \multicolumn{9}{c}{\textbf{0.3052}} \\
				\textbf{R13}: Transformer$^2$ & \multicolumn{9}{c}{\textbf{0.2548}} \\
				\bottomrule
			\end{tabular}			
		}
	\end{center}
	\caption{This table presents empirical findings on Kappa score performance for hybrid text samples, with sub-tables $G0$ to $G3$ showing performances for different text groups categorized by average segment length: $G1$ with $l<3$, $G2$ with $3\leq l\leq 6$, and $G3$ with $l>6$. Enhancements for readability include numbered rows and columns, and \textbf{bolded} values indicating \textbf{top performance} (excluding \texttt{Perfect}) in \textbf{each row}. Note that \texttt{SegFormer} and \texttt{Transformer$^2$} can serve as either \textbf{segment detectors} (results in Column $C6$ and $C7$) or directly as \textbf{segment classifiers} (results in Row $R12$ and $R13$). Each reported entry is a mean over \textbf{three independent runs}.
	} \label{tb:3}
\end{table*}

We present our empirical results in Table \ref{tb:3}. This includes Sub-table $G0$, which reports the Kappa score (a measure of classification accuracy) performance for the entire set of hybrid text samples in the test set. Additionally, Sub-tables $G1$, $G2$, and $G3$ display the performance (Kappa score) for different groups of hybrid texts in the test set. The groups have been categorized based on the average segment length (denoted as $l$) of the hybrid texts, i.e., $l<=3$ in $G1$; $l\in(3,6]$ in $G2$; and $l>6$ in $G3$. This categorization enables our investigation into \textit{whether the choice of the optimal detection strategy changes according to variations in the average segment length of the hybrid texts}. It is noted that a shorter average segment length often indicates more frequent changes in authorship between neighboring sentences, leading to a higher number of boundaries (see Table \ref{tb:tb4}).

\smallskip
\noindent\textbf{Which is the Best Segment Classifier?}
By referring to the $C5$ column (i.e., results of the \texttt{Naive} segment detector) of any one of the sub-tables, we can observe that \texttt{DeBERTa-v3}, \texttt{BERT}, \texttt{SeqXGPT}, \texttt{RoBERTa}, and \texttt{DistilBERT} outperform the other segment classifiers, with \texttt{DeBERTa-v3} being the most effective. Additionally, we notice that the token-classification based methods, namely \texttt{BERT (Token)}, \texttt{DistilBERT (Token)}, and \texttt{RoBERTa (Token)}, consistently underperform compared to their sentence-level classification counterparts, which are \texttt{BERT}, \texttt{DistilBERT}, and \texttt{RoBERTa}, demonstrating the effectiveness of the strategy where a BERT-based encoder is directly combined with a classification layer for sentence-level text classification.

\smallskip
\noindent\textbf{Which is Better? Joint Learning Approaches vs. Two-Step Pipeline Approaches.}
As previously mentioned, \texttt{SegFormer} and \texttt{Transformer$^2$} are approaches that jointly learn the two tasks of segment detection and segment classification and they can serve as either segment detectors or segment classifiers. From our empirical results, we observed an interesting phenomenon: when these two approaches act solely as segment detectors and are combined with high-performing downstream classifiers (e.g., \texttt{DeBERTa-v3}), they may achieve more effective outcomes than when serving directly as segment classifiers. For example, in sub-tables $G2$ and $G3$, the \texttt{SegFormer} (or \texttt{Transformer$^2$}) and \texttt{DeBERTa-v3} pipeline (see $<R1, C6>$ and $<R1, C7>$) outperformed instances where \texttt{SegFormer} (or \texttt{Transformer$^2$}) alone served as the segment classifier ($R12$ and $R13$). These observations suggest that a pipeline with high-performing segment detectors and high-performing segment classifiers does not necessarily perform worse than approaches that jointly learn these two tasks.

\smallskip
\noindent\textbf{How to Choose the Right Strategy? Sentence-by-Sentence Classification vs. Text Segmentation-Based Classification.}
We notice that the best segment detector varies across groups with different average segment length. Specifically, for group $G1$ with an average segment length of $3$ or less, the \texttt{Naive} segment detector achieved the highest performance (averaged across all classifiers), as indicated by a Kappa score of $0.3070$ in $<R11, C5>$; For group $G2$, although the \texttt{Naive} segment detector remains the best in terms of mean performance (see $<R11, C5>$ of the sub-table for $G2$), we observe a notable difference from the results in group $G1$, i.e., only $5$ out of the $10$ top performances (highlighted in \textbf{bold}) originate from column $C5$; Finally, turning to the sub-table for group $G3$, we can see from Row $R11$ that the \texttt{Naive} segment detector performed the worst, while the \texttt{SegFormer} detector achieved the best mean performance.

\begin{figure}[t]
	\centering
	\includegraphics[width=8.1cm, height=3.9cm]{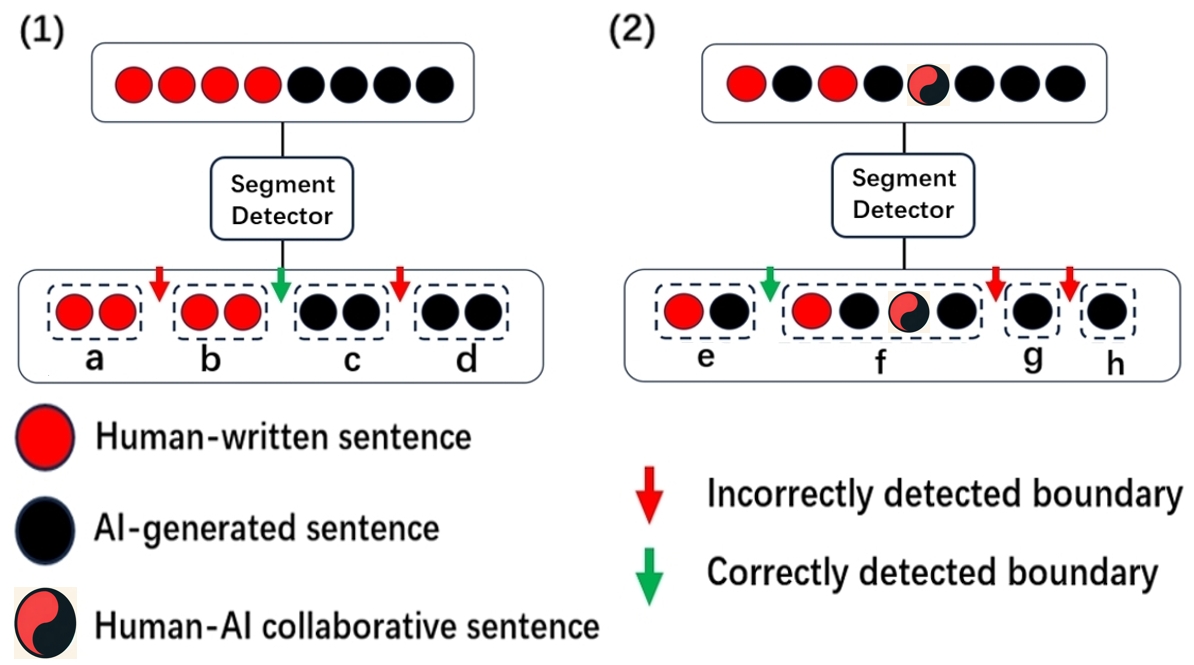}
	\caption{
		This example illustrates the phenomenon that a realistic (imperfect) segment detector is more likely to produce \textbf{authorship-inconsistent segments} (e.g., segments $e$ and $f$) in hybrid texts with a higher number of boundaries than in those with fewer boundaries. Statistics related to this phenomenon can be seen in \textbf{Table \ref{tb:tb4}}.
	} 
	\label{fig:1}
\end{figure}

We first explain why the three realistic segment detectors (i.e., \texttt{SegFormer}, \texttt{Transformer$^2$}, and \texttt{TriBERT}) significantly underperform compared to the \texttt{Naive} segment detector (i.e., sentence-by-sentence classification strategy) in group $G1$ (see sub-table $G1$, Row $R11$). We posit that the reason lies in the fact that hybrid texts in group $G1$ have the shortest average segment length, which typically means a higher number of boundaries (see Table \ref{tb:tb4}). A greater number of boundaries implies increased difficulty in detecting all true boundaries. \textbf{Failing to detect all true boundaries} inevitably leads to the creation of authorship-inconsistent segments, i.e., sentences with differing authorships are incorrectly grouped into a single segment (e.g., segments $e$ and $f$ in Figure \ref{fig:1}) by an imperfect segment detector\footnote{Realistic segment detectors other than \texttt{Perfect} and \texttt{Naive}.}. The sentences within these authorship-inconsistent segments, in theory, cannot be predicted with $100\%$ accuracy by downstream segment classifiers, thereby compromising the effectiveness of sentence-level AI-generated text detection. We provide \textbf{an example in Figure \ref{fig:1}} to illustrate that, given the number of sentences being roughly equal in all hybrid texts (as is the case for the CoAuthor dataset), an imperfect segment detector is more likely to produce authorship-inconsistent segments in hybrid texts with a higher number of boundaries than in those with fewer boundaries: (\romannumeral1) Sub-figure (1) illustrates a hybrid text (with only one true boundary), from which an imperfect segment detector identifies one true boundary and two false boundaries, consequently dividing this text into four segments: $a$, $b$, $c$, and $d$; (\romannumeral2) Sub-figure (2) displays a hybrid text with five boundaries and an imperfect segment detector splits this hybrid text into segments $e$, $f$, $g$, and $h$ based on the three identified boundaries (one is true and the other two are false).
\begin{table}[!htb]
	\begin{center}
		\resizebox{0.44\textwidth}{!}{
			\begin{tabular}{l|c|c|c} 
				\toprule
				\begin{tabular}[c]{@{}c@{}}Average Segment \\Length $l$ within\\ Group \end{tabular}&\begin{tabular}[c]{@{}c@{}}Average Number of \\Boundaries within\\ Group \end{tabular}&\begin{tabular}[c]{@{}c@{}}Ratio of the\\ Group's Samples\\ to the Total \end{tabular} & \multicolumn{1}{c}{\begin{tabular}[c]{@{}c@{}}Ratio of \\Segments with \\Inconsistent \\Authorship \\
					\end{tabular}} \\ 
					\midrule
					G1: $l<=3$  &13.7&46.7\% & 36.30\% \\
					G2: $3<l<=6$  &9.9&27.9\% & 29.70\% \\
					G3: $l>6$  &4.2&25.3\% & 22.90\% \\
					\bottomrule
				\end{tabular}
			}
		\end{center}
		\caption{
			This table details the ratio of segments with inconsistent authorship across different groups of hybrid texts from the test set. 
		} \label{tb:tb4}
	\end{table}
In this example, an imperfect segment detector identified three boundaries in two hybrid texts, each with one true and two false boundaries, resulting in a precision of $1/3$. The recall rates for true boundaries were $1.0$ in the left-side text and $0.2$ in the right-side text, indicating that the detector missed $80\%$ of true boundaries in the right-side text. As mentioned earlier, failing to detect the true boundaries will inevitably result in authorship-inconsistent segments. Take segments $e$ and $f$ in Sub-figure (2) as examples. Due to the mixed-authorship nature of segments $e$ and $f$, if $e$ is predicted to be AI-generated (or human-written), then the human-written (or AI-generated) sentences within this segment will (inevitably and incorrectly) be labeled as AI-generated (or human-written). This is because the downstream classifier will only predict a single authorship for each identified segment. We further define $r$ as the ratio of segments with inconsistent authorship in a hybrid text $ht$ from the test set, using a segment detector (denoted as $dt$) like \texttt{TriBERT}, \texttt{SegFormer}, or \texttt{Transformer$^2$}. We calculate $r$ for each $<ht, dt>$ pair and then average these values within groups $G1$, $G2$, and $G3$ (see Table \ref{tb:tb4}). The results show $G1$ with the highest inconsistency in authorship, followed by $G2$, and $G3$ with the least, supporting our assumptions as seen in Figure \ref{fig:1}. Finally, it should be noted that (\romannumeral1) the \texttt{Perfect} detector identifies all true boundaries without error; (\romannumeral2) the \texttt{Naive} segment detector, which considers each sentence a separate segment, does not miss true boundaries but detects false ones. Interestingly, as shown in Row $R11$ of Sub-Table $G1$, the \texttt{Naive} detector outperforms other detectors, suggesting that the negative impact of false boundaries is less significant than missed boundaries. For example, in Sub-Figure (1), segments $a$ and $b$, originally part of a larger segment $ab$, are split but remain authorship-consistent. Thus, a segment classifier could still theoretically achieve $100\%$ accuracy on $a$ and $b$. However, this unnecessary splitting leads to the challenge of short-text classification \cite{wang2018lstm,guo2023close,wang2021hierarchical}, e.g., predicting separately for $a$ and $b$ tends to be less accurate than predicting for the combined segment $ab$.

The analysis above could also shed light on why the three realistic segment detectors (i.e., \texttt{SegFormer}, \texttt{Transformer$^2$}, and \texttt{TriBERT}) outperform the sentence-by-sentence classification strategy (i.e., \texttt{Naive} segment detector) in group $G3$ (Please refer to Row $R11$, sub-table $G3$, for the exact performance results.). Specifically, we have the following explanation: the average segment length in Group $G3$ is relatively longer and consequently with fewer boundaries, so that (1) the chances of the true boundaries being missed are rarer, resulting in the lower ratio of authorship-inconsistent segments in $G3$ than in $G1$, as can be seen from the far right column of Table \ref{tb:tb4}; and (2) segments with longer length can help alleviate the short-text classification challenge faced by the downstream segment classifier.

\smallskip
\noindent\textbf{Is Sentence-Level AI-generated Text Detection from Hybrid Texts a Challenging Task?}
The results in column $C9$, within sub-table $G0$ (where each cell represents the average performance of a specific segment classifier across all segment detection models), indicate that the performance of all included segment classifiers is generally unsatisfactory. The highest Kappa score recorded is $0.3447$, achieved by \texttt{DeBERTa-v3}. Column $C8$ reveals that pipelines with a \texttt{Perfect} segment detector significantly outperform others, yet the top Kappa score is only $0.5166$, achieved by combining a \texttt{Perfect} segment detector with \texttt{DeBERTa-v3}. 
We posit that the challenging nature of this task can be attributed to the following aspects: (1) firstly, the difficulty could be explained by looking into the human-AI collaboration process \cite{lee2022coauthor} which resulted in the CoAuthor hybrid texts: (1.1) each machine-generated sentence in these hybrid texts was selected (or preferred) by a human writer from a set of suggested sentences, which could cause the chosen generated sentence to deviate from the original distribution of machine-generated sentences, aligning more closely with the distribution of human-written sentences. (1.2) human writers could make edits to the selected sentences, resulting in a third class of sentence (the human-AI collaborative sentence), which is in addition to the purely human-written sentences and the purely AI-generated sentences; (2) secondly, the frequent change of authorship between neighboring sentences within the hybrid text creates difficulties for realistic (imperfect) segment detectors in accurately identifying authorship-consistent segments, compromising the effectiveness of the subsequent authorship identification task, as has been discussed in the previous subsection; (3) lastly, even when the segments are perfectly detected (i.e., Column $C8$), identifying the authorship of these segments remains difficult due to the short length of the segments. As can be seen from Table \ref{tb:tb4}, nearly $50\%$ of the hybrid texts in the test set have an average segment length of $3$ or less.

\section{Conclusion}
This study investigates the recently popular task of AI-generated text detection within human-AI collaborative hybrid texts. Differing from existing studies \cite{seqxgpt_2023,dugan2023real,zeng2023towards} that primarily rely on synthetic and simplistic datasets, we have grounded our research on the CoAuthor dataset \cite{lee2022coauthor}, which encompasses a wide range of diverse and realistic hybrid texts, produced through collaborative efforts between human writers and an intelligent writing system during multi-turn interactions. Although our experimental results suggest that this task is generally challenging, we have distilled some practical tips from our findings to assist potential practitioners and researchers facing this task: (1) with a sufficiently high-performing segment detector and segment classifier, the two-step pipeline that combines text segmentation and segment authorship identification could be more reliable than approaches that jointly learn these two tasks; (2) before embarking on the detection process, it is beneficial to assess (through prior knowledge or other machine learning techniques) the average length of segments (or the number of boundaries) within the hybrid texts. This assessment aids in deciding whether to employ a text segmentation-based strategy for hybrid texts with longer segments (usually having fewer boundaries) or to adopt a direct sentence-by-sentence classification strategy for texts with shorter segments, which often have more boundaries. An application case of the above tips would be to employ a text segmentation-based strategy on hybrid texts known to contain only very few boundaries. This scenario can occur when students aim to spend only the least effort on a writing assignment, e.g., preferring to have a generative LLM finish the assignment from a given start text (resulting in single-boundary hybrid text) rather than engaging in more time-consuming and effort-intensive multi-turn interactions. Moreover, considering that the average segment length of hybrid texts influences the choice of the optimal segment detection strategy, exploring whether the average segment length of a given hybrid text can be accurately predicted represents a promising direction for future research efforts.

\bibliographystyle{named}
\bibliography{ijcai24}
	
\end{document}